\documentclass[preprint,12pt]{elsarticle}

\usepackage{subcaption} % Ensures subfigure is defined
\usepackage{booktabs}
\usepackage{graphicx} % For inserting images

\usepackage{caption}
\usepackage{hyperref}
\usepackage{color}

\usepackage{amssymb}
\usepackage{amsmath}
\journal{journal}

\begin{document}

\begin{frontmatter}



\title{Deformable Attentive Visual Enhancement for Referring Segmentation Using Vision-Language Model} %% Article title


% \author{Alaa Dalaq, Muzammil Behzad} %% Author name

\author[1]{Alaa Dalaq}
% \ead{g202320570@kfupm.edu.sa}

\author[1]{Muzammil Behzad\corref{cor1}}
\ead{muzammil.behzad@kfupm.edu.sa}
\cortext[cor1]{Corresponding author}

\affiliation[1]{organization={King Fahd University of Petroleum and Minerals},%Department and Organization
            % city={},
            country={Saudi Arabia}}

%% Abstract
\begin{abstract}
Image segmentation is a fundamental task in computer vision, aimed at partitioning an image into semantically meaningful regions. Referring image segmentation extends this task by using natural language expressions to localize specific objects, requiring effective integration of visual and linguistic information. In this work, we propose SegVLM, a vision-language model that incorporates architectural improvements to enhance segmentation accuracy and cross-modal alignment. The model integrates squeeze-and-excitation (SE) blocks for dynamic feature recalibration, deformable convolutions for geometric adaptability, and residual connections for deep feature learning. We also introduce a novel referring-aware fusion (RAF) loss that balances region-level alignment, boundary precision, and class imbalance. Extensive experiments and ablation studies demonstrate that each component contributes to consistent performance improvements. SegVLM also shows strong generalization across diverse datasets and referring expression scenarios.
\end{abstract}


% \begin{highlights}
% \end{highlights}

%% Keywords

\end{frontmatter}

\section{Introduction}
\label{sec:intro}

Image segmentation is a key computer vision task that involves dividing an image into meaningful regions based on their semantics, such as objects, textures, or scene boundaries \cite{long2015fcn}. Segmentation enables pixel-level classification, going beyond the coarser detections provided by traditional object detection methods. The high pixel-level accuracy of segmentation proves valuable in numerous real-world applications. In the medical field, it aids in precisely locating tumors, lesions, and anatomical structures, supporting accurate diagnosis and strategic planning. \cite{ma2024medsam}. In autonomous driving, segmentation aims to distinguish roads, pedestrians, and obstacles to guide it along a safe path \cite{hao2024real}. Other important applications involves satellite imaging \cite{yu2024semantic}, and systems based on analogous reality where the boundaries of objects must be exactly followed \cite{zhang2024augmented}. These diverse and demanding applications have driven significant improvements in the performance of segmentation algorithms \cite{chen2017deeplab}.\\ Recent advancements in deep learning, in particular convolutional neural networks (CNNs) \cite{liu2023gres, yan2023mmnet}, encoder-decoder architectures \cite{li2021mail, ding2021vision} and transformer-based models \cite{cho2024metris, xiao2024oneref}, have led to great improvements in segmentation performance and are now fundamental elements of modern visual systems. Furthermore, the technological advances presented in the literature have laid the groundwork for more complex multimodal segmentation systems \cite{reza2023mmsformer}. More recently vision-language models (VLMs) have pushed boundaries of image segmentation by closely combining the understanding of natural language with visual perception \cite{radford2021learning}. These models enable tasks like referring expression segmentation, which involves segmenting an object from an image based on a natural language description \cite{hu2016segmentation}. By linking language with vision, VLMs offer a compelling approach to aligning language with perception, enabling natural and intuitive human-AI interactions. \cite{ruan2023seem}. 

More recently, transformer-based VLM models, like contrastive language image pre-training (CLIP) \cite{radford2021learning} and bootstrapping language-image pre-training (BLIP) \cite{li2022blip} have obtained impressive gains in vision-language pretraining, leading to zero-shot and few-shot segmentation with little human supervision \cite{li2022blip}. These models are applied to more complex applications, including robot perception, visual question-answering, and assistive technology, where user's intent with context is crucial. Besides, recent models like SEEM \cite{ruan2023seem} show demonstrate that integrating various segmentation tasks under a unified vision-language framework can surpass task-specific models in cross-dataset generalization. Therefore, VLMs are becoming increasingly vital in building segmentation systems that are not only accurate but also flexible, interpretable, and aligned with human understanding.

\subsection{Motivations}
Referring expression segmentation aims to find instances of objects in an image that match an expression provided in natural language, thus requiring the model to jointly parse visual and linguistic inputs \cite{liu2023gres}. Recent methods, including the nearby Cross-modal Referring Image Segmentation (CRIS) model \cite{wang2022cris}, have achieved impressive performance on popular benchmarks including RefCOCO \cite{yu2016modeling}, RefCOCO+ \cite{yu2016modeling}, and RefCOCOg \cite{mao2016generation}. However, these datasets are quite small in terms of linguistic diversity and object complexity. One more challenging datasets like PhraseCut \cite{wu2020phrasecut}, where referring expressions are longer and more compositional and object distinctions are finer, the CRIS \cite{wang2022cris} model fails to produce accurate segmentation. This motivates improvement of model generalization to linguistically and visually complex situations.

To mitigate these limitations, we present a set of architectural and that improves the segmentation performance. The goal is to enhance the segmentation capability on the PhraseCut dataset by focusing visual-linguistic alignment and handling diverse and ambiguous references effectively. These advancements have practical value in areas like like human-computer interaction \cite{liu2023referring, alalyani2024multimodal, alalyani2024scmre}, robotics \cite{Tang_2023_CVPR,shridhar2018interactive, paul2020grounding} and assistive technologies where accurately interpreting natural language and segmenting objects is critically important \cite{yu2018mattnet}.
\begin{figure}[t!]
    \centering
    \includegraphics[width=0.8\linewidth]{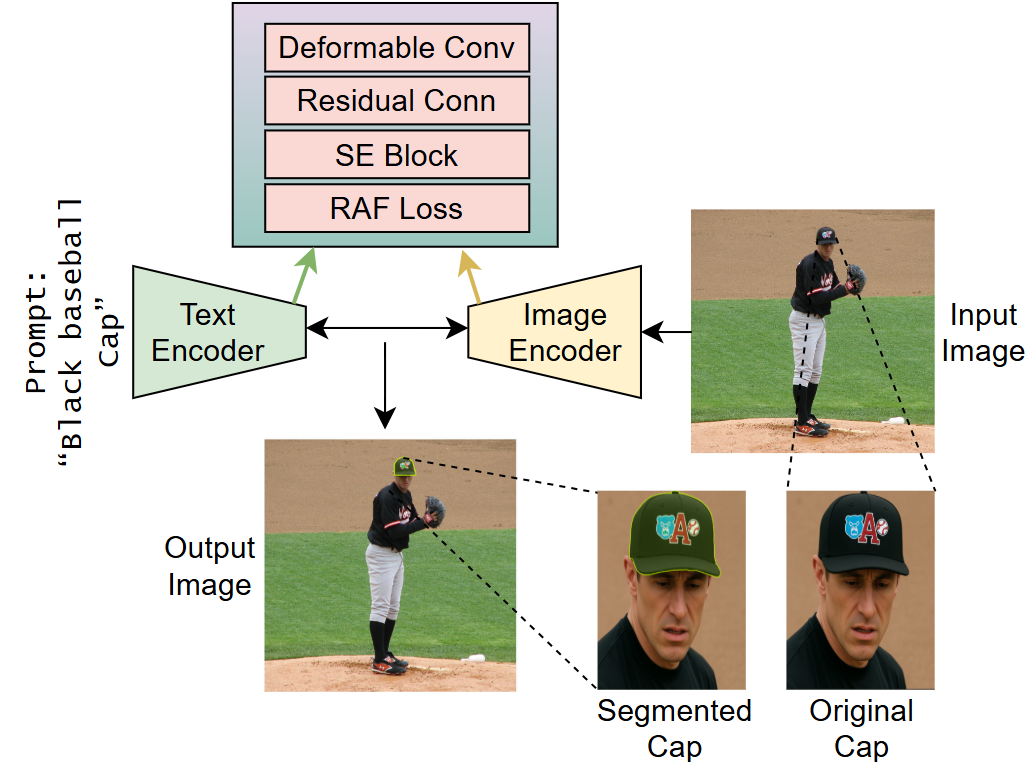}
    \caption{
        Overview of SegVLM Vision-Language Segmentation. Given the input image and a referring expression (e.g., “Black baseball cap”), SegVLM outputs a segmented mask for the relevant region. 
        The model integrates our proposed novelties such as RAF Loss, Deformable Convolution, Residual Connections, and SE Attention to refine segmentation performance.
    }
    \label{fig:cris_overviewA}
\end{figure}

\subsection{Contributions}
We use CRIS \cite{wang2022cris} as our baseline and extend it to propose SegVLM: a VLM-based segmentation model as shown in Fig. \ref{fig:cris_overviewA}. SegVLM addresses the above-mentioned limitations in handling more complex referring expression segmentation tasks.

\begin{itemize}
  \item We implement architectural modifications to enhance the visual-linguistic integration process, enabling the model to better align image features with complex and compositional referring expressions. These changes aim to improve the model’s internal representational capacity while keeping the design lightweight and practical.
  
  \item We introduce RAF Loss, a hybrid objective combining Binary Cross-Entropy, Dice, and Focal loss to address class imbalance, penalize boundary errors, and emphasize hard-to-classify pixels in the segmentation masks.
  
  \item We integrate deformable convolutional layers to allow adaptive receptive fields that dynamically focus on irregular object boundaries, improving spatial localization of referred regions.
  
  \item We embed Squeeze-and-Excitation (SE) blocks to adaptively recalibrate channel-wise feature responses, reinforcing discriminative visual cues conditioned by linguistic context.
\end{itemize}
This approach solves two fundamental problems in segmentation tasks: boosting the accuracy of detecting object borders while also dealing with class imbalance by lowering the impact of easier classes like the background and large objects on the network training. Thanks to these improvements, we achieve a significant gain in segmentation quality on PhraseCut \cite{wu2020phrasecut}, a dataset noted for its complex natural language queries and high diversity of appearances of the objects. Importantly, all modifications are computationally efficient, ensuring that our proposed SegVLM model retains lower number parameters footprint and remains suitable for real-time use and deployment in resource-constrained environments. Our contributions advance interactive and language-guided segmentation by improving robustness, accuracy, and usability, all while maintaining performance and efficiency.

\section{Related Work}
\label{sec:format}

\subsection{Vision-Language Models}
Over the past few years, vision-language models (VLM) has made substantial progress, by significantly improving performance across various multi-modal tasks. Modern approaches leverage large-scale image-text pairs from extensive image datasets to jointly learn aligned representations across visual and language modalities. Several models \cite{NEURIPS2023_6dcf277e,NEURIPS2024_9ee3a664,NEURIPS2024_03738e5f,pmlr-v139-kim21k,pmlr-v139-radford21a,yu2016modeling,yang2022unitab,wang2022image,bai2023qwen} extend visual language processing by integrating vision encoders and instruction-tuned language models, facilitating functions such as image captioning, visual question answering, and grounded reasoning. In this context BLIP-2 \cite{pmlr-v202-li23q} developed a streamlined querying transformer to effectively connect static visual encoders with massive language models. Simultaneously, MiniGPT-4 \cite{zhu2024minigpt} exhibited remarkable visual comprehension using a pre-trained backbone refined with paired instructional data. Recent studies have expanded the foundational success of CLIP-style contrastive learning \cite{radford2021clip} by adapting VLM models for complex tasks such as referring expression segmentation \cite{10.1145/3581783.3612117,lads2023aaai,hemanthage2024recantformer}, video comprehension \cite{yang2023videococa,mei2024slvp,zou2023xdecoder,wang2022omnivl}, and multi-modal generation \cite{dai2023instructblip}, demonstrating that transferring high-level alignment knowledge can significantly improve downstream applications.

\subsection{Contrastive Learning}
Initial studies on contrastive learning \cite{1640964} laid the groundwork by utilizing comparisons between positive and negative pairings to acquire discriminative representations.  Subsequent improvements \cite{pmlr-v119-chen20j,10.1007/978-3-031-19809-0_17,he2020moco,Li_2022_CVPR,10.1109/TMM.2023.3324588,10.1145/3581783.3612247} enhanced this concept by considering each image instance as a distinct class and maximizing instance-level differentiation using contrastive loss.  To apply these paradigms to dense prediction challenges, models such as VADeR \cite{pinheiro2020vader} and DenseCL \cite{wang2021dense} investigated pixel-wise contrastive learning, facilitating enhanced feature alignment at the spatial level.  Advancing beyond single-modal contexts, CLIP \cite{radford2021clip} established a robust cross-modal contrastive framework that acquires transferable representations through the alignment of extensive image-text pairs.  This approach has subsequently inspired various vision-language segmentation algorithms, like CRIS \cite{wang2022cris}, which adeptly applies knowledge from CLIP to referencing image segmentation. The architecture of CRIS \cite{wang2022cris} facilitates comprehensive segmentation directed by natural language questions, illustrating the application of high-level vision-language alignment for pixel-level comprehension.

\subsection{Referring Image Segmentation}

Referring image segmentation, which was initially highlighted in \cite{hu2016segmentation}, aims at locating a particular area in an image that correlates to a given natural language statement.  Initial techniques \cite{hu2016segmentation, mao2016generation,li2018referring,liu2017recurrent, margffoy2018dynamic}often utilized CNNs for visual feature extraction and long-short-term memory (LSTMs) \cite{hochreiter1997long} to encode referring phrases, integrating these modalities through feature concatenation and producing segmentation masks using fully convolutional networks (FCNs) \cite{long2015fully}.  Subsequently, two-stage architectures utilizes \cite{yu2018mattnet}, instance segmentation models such as Mask R-CNN \cite{he2017mask} to provide object recommendations, succeeded by language-guided selection of the target instance.  The modular co-attention network (MCN) \cite{luo2020multi} accelerated the research in this direction by concurrently learning segmentation and expression inside a robust framework. Similarly, CRIS \cite{wang2022cris} developed a revolutionary methodology that utilizes the robust vision-language alignment originally proposed by CLIP \cite{radford2021clip} to direct segmentation in a comprehensive manner.  By leveraging cross-modal knowledge from CLIP \cite{radford2021clip}, CRIS \cite{wang2022cris} attains strong performance in reference segmentation tasks without the necessity for challenging multi-stage pipelines, signifying a notable advancement in vision-language pretraining-based segmentation frameworks.

More importantly, attention mechanisms have been essential in enhancing referring picture segmentation by facilitating precise matching between visual and language signals.  Numerous research have investigated various attention-based ways to improve cross-modal reasoning.  For example, the studies \cite{zhang2023risam,yang2022lavt, ouyang2023slvit} have presented a vision-guided linguistic attention model that adaptively enhances sentence context according to the content of visual regions. The study \cite{ye2019cross} introduced a cross-modal self-attention (CSMA) module that simultaneously emphasizes prominent visual and textual characteristics to enhance mutual comprehension. Building on prior vision-language methods, the authors in \cite{hu2020bi} proposed a bi-directional attention mechanism that enables mutual guiding between language and visual through specialized attention modules, similarly several studies have used bi-directional cross-modal attention to improve referring picture segmentation. RISAM \cite{zhang2023risam} introduces mutual-aware attention features, whereas Fuse and Calibrate \cite{yan2024fuse} and Prompt-Guided Fusion \cite{shang2024prompt} refine vision-language interaction via guided and prompt-based attention flows.  In contrast to the traditional methodologies, most recent contemporary models, most recent contemporary models like CRIS \cite{wang2022cris} employ pretrained cross-modal frameworks to enhance segmentation through improved visual-linguistic alignment.  By integrating extensive vision-language knowledge into the segmentation pipeline, such architectures present a promising alternative to traditional attention techniques, enhancing the model's capacity for precise cross-modal matching.

\section{SegVLM: A Vision-Language Segmentation Model}
This section introduces our modified model, which generates segmentation masks based on user-provided natural language descriptions, rather than relying on spatial cues such as extreme points or bounding boxes \cite{yu2018mattnet}.We first present the model architecture, followed by our approach for integrating user-provided natural language expressions into the segmentation pipeline. Finally, we detail the training strategy, which complements the architectural modifications by introducing a novel loss function alongside additional techniques to improve segmentation performance.
\begin{figure*}[t!]
    \centering
    \includegraphics[width=\textwidth]{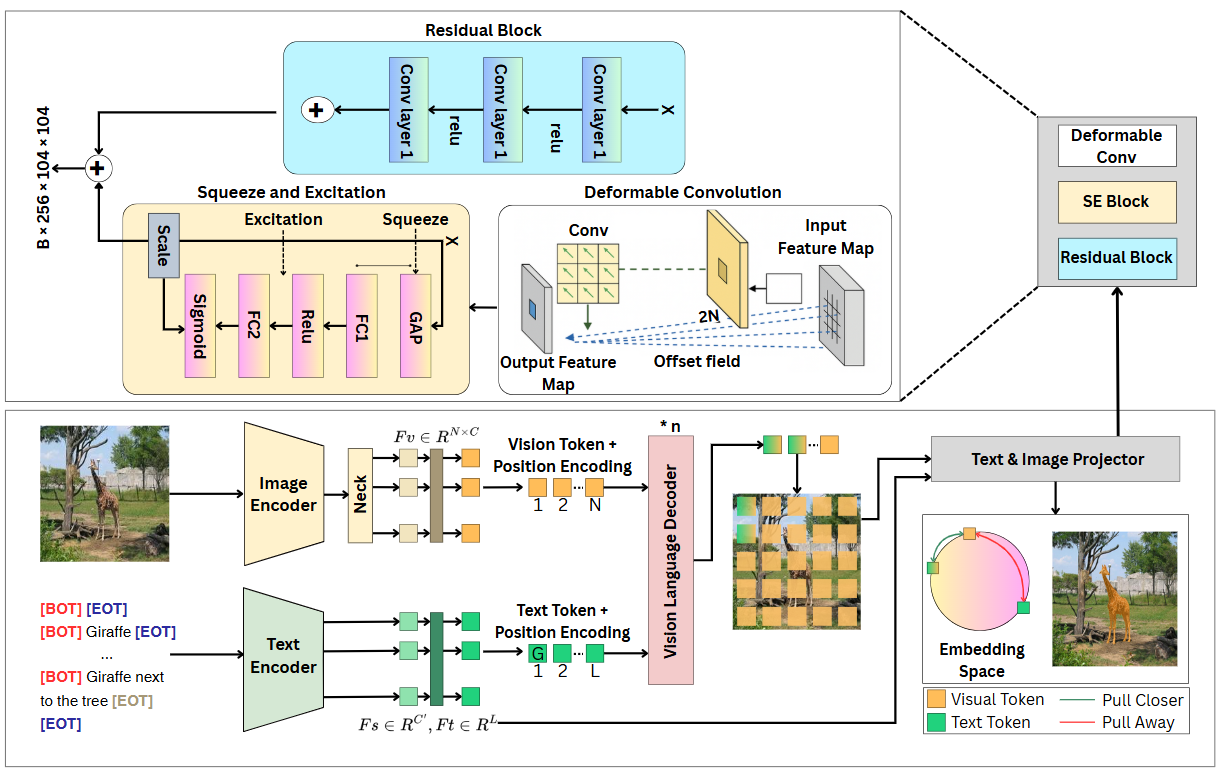}
    \caption{The SegVLM architecture for referring image segmentation comprises an image encoder and a text encoder, whose outputs are fused by a vision-language decoder operating on visual and textual tokens with positional encoding. The fused cross-modal features are then refined using a Text \& Image Projector, which aligns both modalities within a shared embedding space. To further enhance the visual feature representation, a Residual Block is incorporated, augmented with Squeeze-and-Excitation (SE) blocks and Deformable Convolutions. These additions enable adaptive channel recalibration and flexible spatial context modeling, collectively contributing to improved object localization and robust cross-modal grounding for complex referring expressions.}
    \label{fig:cris_overview}
\end{figure*}

\subsection{Model Architecture}
We first apply an enhancement block in the text and image projector to improve the image and text feature fusion. This block comprises deformable convolution layers, a Squeeze-and-Excitation (SE) module, and residual shortcut connections, facilitating attention to spatial and semantic cues adaptively. Second, we propose a novel loss function to treat class imbalance better and increase sensitivity to challenging samples.

In Fig. \ref{fig:cris_overview}, we show an overview of our proposed SegVLM model architecture. As illustrated, the multi-modal projection begins when the input image and referring expression are initially encoded by the baseline model. Afterwards, these refined features are subsequently transformed into a segmentation mask. With this formulation, our model is able to learn complex language and vision associations that allow it to produce semantically relevant and consistent segmentation.

\subsection{Multimodal Projection for Semantic and Spatial Alignment}
To solve the precise vision-language alignment challenge for referring image segmentation, we introduce a refined multimodal projection module framework. Unlike the static transformations provided by the original projection layers of our baseline model on fused features, our design contains two adaptive mechanisms that increase spatial and semantic alignment. The salient features of this module are as follows:
\begin{itemize}
    \item A spatially-aware visual enhancement block that adaptively refines image features using deformable convolution, residual learning, and attention mechanisms; and,
    \item A language-modulated dynamic convolution module that generates instance-specific convolution kernels conditioned on the input referring expression.
\end{itemize} 
Together, these modules enable the model to capture the highly complex, context-dependent relationships between natural language and visual content, ultimately producing more accurate and expressive segmentation masks.

\subsection{Deformable-Attentive Visual Enhancement}
The adaptive projection module enhances the fused visual features by learning region-level attention that is both semantically and spatially relevant to the input image. Starting from a feature map $x \in \mathbb{R}^{B \times C \times H \times W}$. we first perform a deformable convolution, which adds spatial flexibility in standard convolution with learnable offsets delta-p as follows:

\begin{equation}
\Delta \mathbf{p} = \mathcal{F}_{\text{offset}}(\mathbf{x}), \quad
\mathbf{x}_{\text{dcf}} = \boldsymbol{\phi}(\mathbf{x}, \Delta \mathbf{p}),
\end{equation} where $\phi$ denote the deformable convolution operation. By applying this operation, the receptive field can dynamically adapt to variations in object shapes, which has proven effective in dense prediction tasks \cite{dai2017deformable}, as the fixed-grid kernels may not satisfy the diverse learning complexity.
Then, we follow this with a Squeeze-and-Excitation block to re-calibrate the features channel-wise as:

\begin{equation}
\mathbf{s} = \sigma\left( \mathbf{W}_2 \cdot \delta\left( \mathbf{W}_1 \cdot \text{GAP}(\mathbf{x}_{\text{dcf}}) \right) \right), \quad
\mathbf{x}_{\text{sc}} = \mathbf{s} \odot \mathbf{x}_{\text{dcf}},
\end{equation}where $\text{GAP}$ denotes global average pooling, $\delta$ is the ReLU activation, $\sigma$ is the sigmoid function, and $\odot$ denotes element-wise multiplication. This approach selectively emphasizes informative channels for spatially-aware segmentation performance. Finally, in order to retain lower-level spatial structure,we introduce a residual shortcut connection that is upsampled and passed through a lightweight convolutional transformation:

\begin{equation}
\mathbf{x}_{\text{res}} = \mathcal{F}_{\text{residual}}\left(\mathcal{U}(\mathbf{x})\right), \quad
\mathbf{x}_{\text{fused}} = \mathbf{x}_{\text{se}} + \mathbf{x}_{\text{res}}.
\end{equation}
This shortcut approach enhances the final representation by conserving fine-grained spatial information, which in turn makes it easier to achieve more accurate boundary localization and overall segmentation performance.

\subsection{Referring-Aware Fusion Loss}
We propose a novel Referring-Aware Fusion (RAF) Loss to address the challenges of class imbalance and hard region segmentation in referring image segmentation tasks. We formulate this loss to effectively balance easy and hard pixel predictions, emphasizing misclassified and boundary pixels, thereby guiding the model toward more precise and resilient segmentation outcomes by integrating Binary Cross-Entropy (BCE) \cite{goodfellow2016deep}, Focal Loss \cite{lin2017focal}, and a pixel-wise Adaptive Dice Loss \cite{milletari2016vnet}. In this regard, let $p \in [0, 1]$ be the predicted probability and $y \in \{0, 1\}$ be the ground truth label. The standard BCE loss is given by:

\begin{equation}
\mathcal{L}_{\text{BCE}} = - \sum_i \left[ y_i \log(p_i) + (1 - y_i) \log(1 - p_i) \right].
\end{equation}
However, BCE is often swamped by easy negatives in class-imbalanced settings. To mitigate this, we adopt a modulated version by incorporating Focal Loss, which down-weights easy examples and concentrates the training on harder, misclassified instances. This is expressed as follows:

\begin{align}
\mathcal{L}_{\text{focal}} = - \sum_i \Big[
    &\alpha \, y_i \, (1 - p_i)^\gamma \log(p_i) \notag \\
    &+ (1 - \alpha)(1 - y_i) \, p_i^\gamma \log(1 - p_i)
\Big],
\label{eq:focal}
\end{align}where $\alpha \in (0, 1)$ balances positive vs. negative samples, and $\gamma > 0$ is the focusing parameter. Although BCE \cite{goodfellow2016deep} and Focal Loss \cite{lin2017focal} are pixel-wise methods, they do not explicitly consider the overlap between predicted and ground-truth masks. To address this, we add an pixel-wise adaptive Dice Loss \cite{milletari2016vnet}, which measures set-level similarity as expressed as below:

\begin{equation}
\mathcal{L}_{\text{AdaptiveDice}} = 1 - \frac{2 \sum_i \alpha_i p_i y_i + \epsilon}{\sum_i \alpha_i p_i + \sum_i \alpha_i y_i + \epsilon}.
\end{equation}
Here, $p_i$ and $y_i$ denote the predicted probability and ground truth at pixel $i$, respectively. The pixel-wise weight $\alpha_i$ can be defined as:

\begin{equation}
\alpha_i = |p_i - y_i| \quad \text{or} \quad \alpha_i = (1 - p_i)^\gamma.
\end{equation}
This helps focus more on uncertain or misclassified pixels, where E($\epsilon$) is a small constant for numerical stability.
The overall RAF Loss is then formulated as:

\begin{align}
\mathcal{L}_{\text{RAF}} = & - \lambda_1 \sum_{i} \left[ y_i \log(p_i) + (1 - y_i) \log(1 - p_i) \right] \nonumber \\
& - \lambda_2 \sum_{i} \Big[ \alpha \, y_i (1 - p_i)^{\gamma} \log(p_i) \nonumber \\
& \qquad\qquad + (1 - \alpha)(1 - y_i) \, p_i^{\gamma} \log(1 - p_i) \Big] \nonumber \\
& + \lambda_3 \left( 
1 - \frac{2 \sum_{i} \alpha_i p_i y_i + \epsilon}
{\sum_{i} \alpha_i p_i + \sum_{i} \alpha_i y_i + \epsilon} 
\right),
\end{align}
where $\lambda_1$, $\lambda_2$, and $\lambda_3$ control the contribution of each term. This formulation allows our model to to allow context-aware segmentation, while remaining robust to class imbalance and soft boundaries. In the subsequent sections, we empirically demonstrate that this fusion achieves superior performance in terms of mask quality and training stability compared to any of the individual loss components.

\section{Experimental Details}
\subsection{Dataset and preprocessing}
The experiments in this paper are conducted on the PhraseCut dataset \cite{wu2020phrasecut}, a large-scale benchmark specifically for the referring image segmentation task. It consists of more than 340k region-phrase pairs from the Visual Genome dataset. To allow efficient data access and large-scale training, we convert the dataset into lightening memory-mapped database (LMDB) \cite{chu2019lmdb} and aligned image-mask expression triplets. We tokenize all referring expressions with our model's tokenizer and truncate or pad to a maximum length. The ground-truth segmentation masks are binarized and resized to fit the model input resolution.

\subsection{Model Configurations and Training Protocols}
Referring expressions are encoded into 512-dimensional embeddings using the text encoder. For the visual feature extraction, we use the ResNet-101 backbone to generate 512-dimensional spatial feature maps. This common dimensionality supports multi-modal fusion in the projection and decoding modules, the model is trained for 50 epochs on batches of size 64. We use the Adam optimizer with an initial learning rate of $1 \times 10^{-4}$ where we decay the learning rate at epochs 15 and 30 with a decay factor of 0.1. We also utilize synchronized Batch Normalization \cite{zhang2018context} on multiple GPUs and gradient clipping \cite{pascanu2013difficulty} with clipnorm set to 0 (prevents exploding gradients) for lower loss function sensitivity. All convolutional layers are initialized by Kaiming normalization \cite{he2015delving}, while bilinear interpolation is used for upsampling.
\subsection{Evaluation and Runtime Details}
To provide a complete measure of segmentation quality, we evaluate model performance on the test split of PhraseCut on various metrics. Besides the standard mean Intersection-over-Union (mIoU) metric for evaluating the overall overlap between the predicted and ground-truth masks, we present precision scores with several Intersection-over-Union thresholds: Prec@50, Prec@60, Prec@70, Prec@80 and Prec@90. IoU metrics indicate the proportion of predictions that exceed the corresponding IoU threshold with the ground truth, providing an indication of segmentation quality across varying levels of strictness. The experiments were carried on a distibuted NVIDIA GeForce RTX 3090. The model took approximately 21 hours for training. During inference, the model achieves an average of 40-45 FPS with input dimensions of 416 × 416.

\section{Results \& Discussions}
\label{sec:format}
\subsection{Quantitative Results}
We evaluate the effectiveness of our proposed SegVLM model on the PhraseCut test set. The evaluation includes the standard Intersection-over-Union (IoU) metric as well as precision at multiple IoU thresholds (Prec@50, Prec@60, Prec@70, Prec@80, and Prec@90), offering a comprehensive view of both segmentation quality and localization accuracy.
Additionally, we compare our model against several state-of-the-art benchmarks including MattNet \cite{yu2018mattnet}, MDETR ENB3 \cite{Kamath_2021_ICCV}, HULANet \cite{wu2020phrasecut}, RMI \cite{liu2017recurrent}, and GROUNDHOG~\cite{zhang2024groundhog}. Following this, we also conduct ablation studies to analyze the individual contributions of each enhancement including the proposed RAF Loss, the deformable convolution with residual connection, and the SE block integrated into the projector module. Finally, we present graphical visualizations that highlight performance trends across thresholds and variants.
\begin{table*}[t!]
\centering
\footnotesize
\caption{Quantitative comparison with state-of-the-art methods on the PhraseCut dataset. SegVLM outperforms all baselines across precision metrics, especially at higher IoU thresholds, while maintaining strong overall IoU performance.}
\resizebox{\linewidth}{!}{%
\begin{tabular}{lcccc}
\toprule
\textbf{Model} & \textbf{IoU (\%)} (\color{blue} $\uparrow$\color{red} $\downarrow$) & \textbf{Prec@50} (\color{blue} $\uparrow$\color{red} $\downarrow$) & \textbf{Prec@70} (\color{blue} $\uparrow$\color{red} $\downarrow$) & \textbf{Prec@90} (\color{blue} $\uparrow$\color{red} $\downarrow$) \\
\midrule
GROUNDHOG \cite{Zhang_2024_CVPR}      & 54.50 (\color{red}0.63$\downarrow$)   & --     & --     & --     \\
MDETR ENB3 \cite{Kamath_2021_ICCV}     & 53.70 (\color{blue}0.17$\uparrow$)   & 57.50 (\color{blue}0.73$\uparrow$)   & 39.90  (\color{blue}0.91$\uparrow$)  & 11.90 (\color{blue}0.74$\uparrow$)   \\
HULANet \cite{wu2020phrasecut}        & 41.30 (\color{blue}12.57$\uparrow$)   & 42.90  (\color{blue}15.33$\uparrow$)  & 27.80  (\color{blue}13.01$\uparrow$) & 5.90 (\color{blue}6.74$\uparrow$)     \\
RMI \cite{liu2017recurrent}           & 21.10 (\color{blue}32.77$\uparrow$)   & 22.00 (\color{blue}36.23$\uparrow$)   & 11.60 (\color{blue}29.21$\uparrow$) & 1.50 (\color{blue}11.14$\uparrow$)     \\
MattNet \cite{yu2018mattnet}          & 20.20 (\color{blue}33.67$\uparrow$)   & 19.70 (\color{blue}38.53$\uparrow$)   & 13.50  (\color{blue}27.31$\uparrow$) & 3.00  (\color{blue}9.64$\uparrow$)   \\
\textbf{SegVLM (Ours)}                & \textbf{53.87} & \textbf{58.23} & \textbf{40.81} & \textbf{12.64} \\
\bottomrule
\end{tabular}
}
\label{tab:benchmark1}
\end{table*}
\begin{table*}[b!]
\centering
\footnotesize
\caption{Ablation study on the PhraseCut dataset, showing incremental improvements over the baseline, with our proposed SegVLM model achieving the best overall performance.}
\resizebox{\linewidth}{!}{%
\begin{tabular}{lcccccc}
\toprule
\textbf{Method Variant} & \textbf{IoU} (\color{blue} $\uparrow$\color{red} $\downarrow$) & \textbf{P@50} (\color{blue} $\uparrow$\color{red} $\downarrow$) & \textbf{P@60} (\color{blue} $\uparrow$\color{red} $\downarrow$) & \textbf{P@70} (\color{blue} $\uparrow$\color{red} $\downarrow$) & \textbf{P@80} (\color{blue} $\uparrow$\color{red} $\downarrow$) & \textbf{P@90} (\color{blue} $\uparrow$\color{red} $\downarrow$) \\
\midrule
Baseline         & 43.57 (\color{blue}10.30$\uparrow$)  & 45.10 (\color{blue}12.00$\uparrow$) & 35.40 (\color{blue}9.70$\uparrow$)  & 25.10 (\color{blue}10.90$\uparrow$)  & 14.20 (\color{blue}8.60$\uparrow$) & 4.10 (\color{blue}6.20$\uparrow$)  \\
+ RAF Loss                               & 48.25 (\color{blue}5.62$\uparrow$)  & 50.70 (\color{blue}6.40$\uparrow$)  & 39.80 (\color{blue}5.30$\uparrow$)  & 29.60 (\color{blue}6.40$\uparrow$) & 17.10 (\color{blue}5.70$\uparrow$) & 6.50 (\color{blue}3.80$\uparrow$)  \\
+ Deformable + Residual
& 51.34 (\color{blue}2.53$\uparrow$)  & 54.30 (\color{blue}2.80$\uparrow$)  & 42.70 (\color{blue}2.40$\uparrow$) & 33.20 (\color{blue}2.80$\uparrow$) & 20.20 (\color{blue}2.60$\uparrow$) & 8.20 (\color{blue}2.1$\uparrow$)  \\
+ SE Block (SegVLM)           & \textbf{53.87} & \textbf{57.10} & \textbf{45.10} & \textbf{36.00} & \textbf{22.80} & \textbf{10.30} \\
\bottomrule
\end{tabular}%
}
\label{tab:ablation2}
\end{table*}

In Table \ref{tab:benchmark1}, we compare SegVLM with recent state-of-the-art models on the PhraseCut. It can be observed that our model achieves an IoU score of 53.87\%, outperforming MDETR ENB3 (53.7\%) and significantly surpassing earlier methods such as HULANet (41.3\%), RMI (21.1\%), and MattNet (20.2\%). Across all mean average precision metrics at different IoU thresholds, SegVLM consistently outperforms all baselines, achieving 58.2\% at Prec@50, 40.8\% at Prec@70, and 12.6\% at Prec@90. These results are consistent with our architectural improvements and the proposed RAF loss, indicating that our model not only enhances overall segmentation overlap but also produces more accurate mask boundaries, as evidenced by the gains at higher IoU thresholds (e.g., Prec@90). SegVLM delivers strong performance across all evaluation metrics, despite its relatively simple architecture and moderate size of approximately 63 million parameters. This is especially noteworthy when compared to more complex and heavier models such as GROUNDHOG and MattNet, which require significantly more computational resources.

\begin{figure}[t!]
    \centering
    \includegraphics[width=1.0\linewidth]{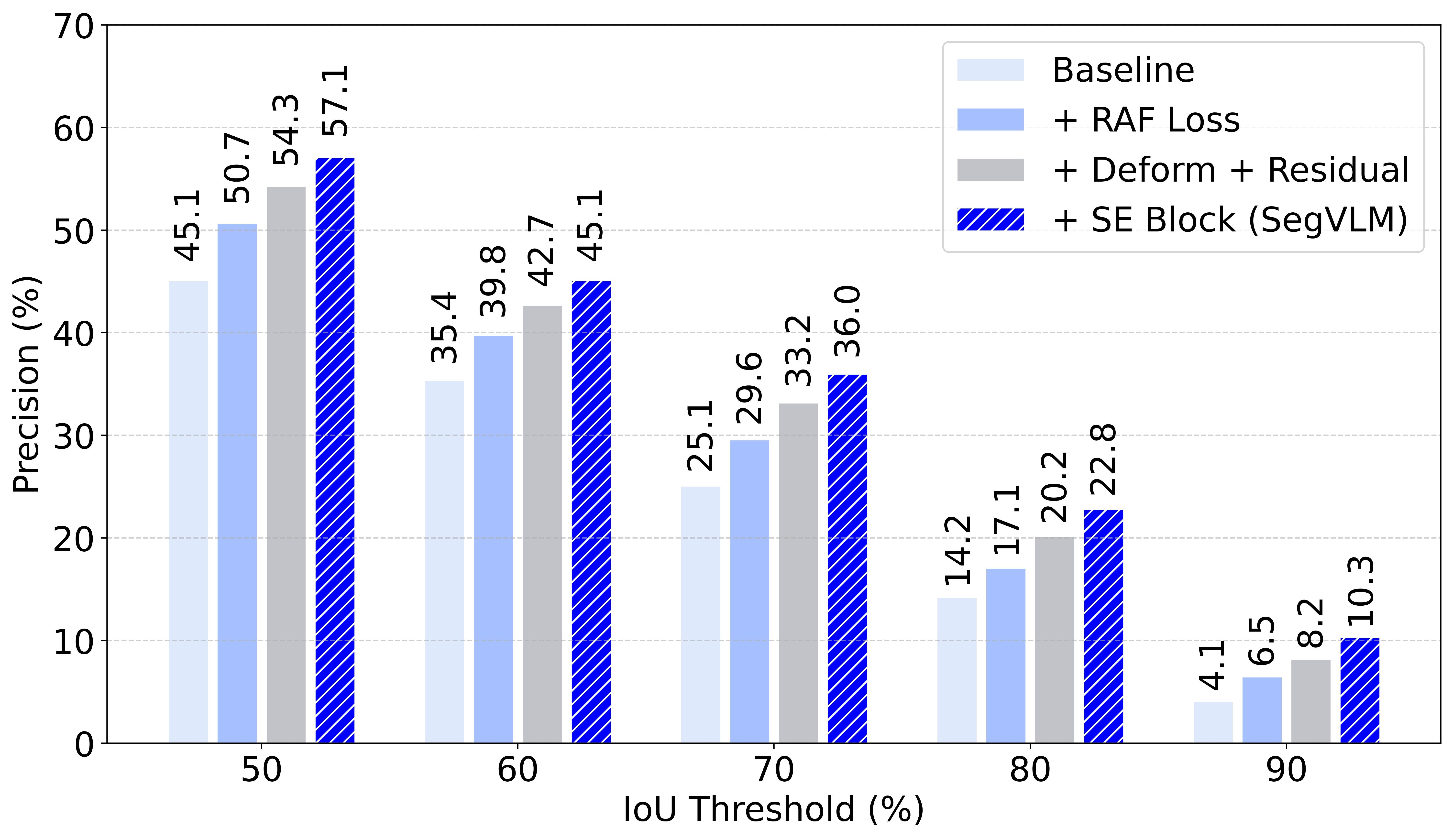}
    \caption{Ablation study showing the impact of each proposed enhancement on IoU, Prec@50, and Prec@90. Each component contributes incrementally to the final performance of SegVLM, with consistent improvements across all metrics, especially at stricter thresholds (Prec@90).}
    \label{fig:ablation1}
\end{figure}
\begin{figure}[t!]
    \centering
    \includegraphics[width=1.0\linewidth]{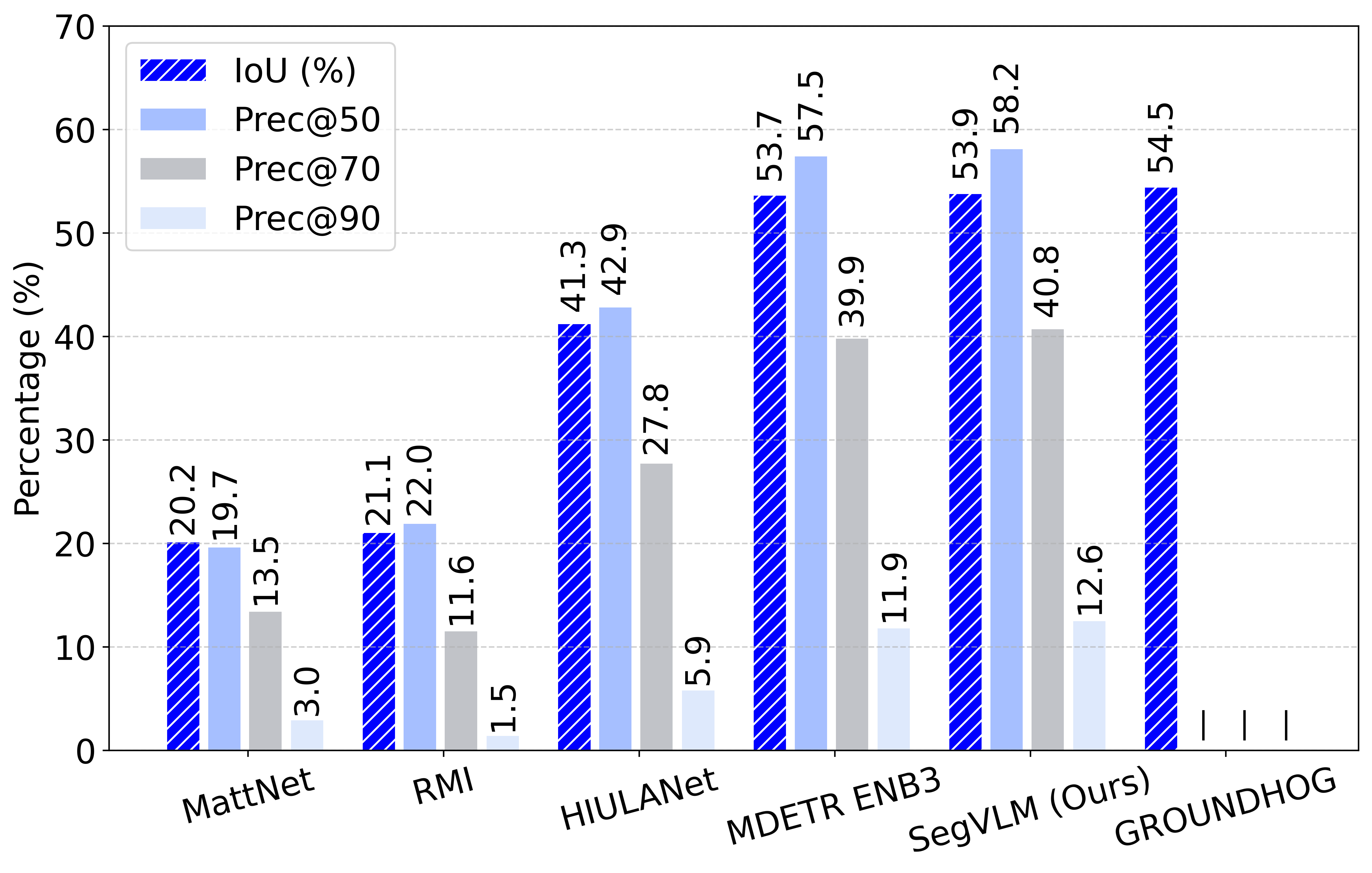}
    \caption{Precision across IoU thresholds (50–90). SegVLM consistently achieves higher precision at all thresholds, with particularly notable gains at stricter levels (Prec@80 and Prec@90), indicating improved boundary alignment and overall robustness.}
    \label{fig:ablation2}
\end{figure}
\begin{figure}[t!]
    \centering
    \includegraphics[width=1.0\linewidth]{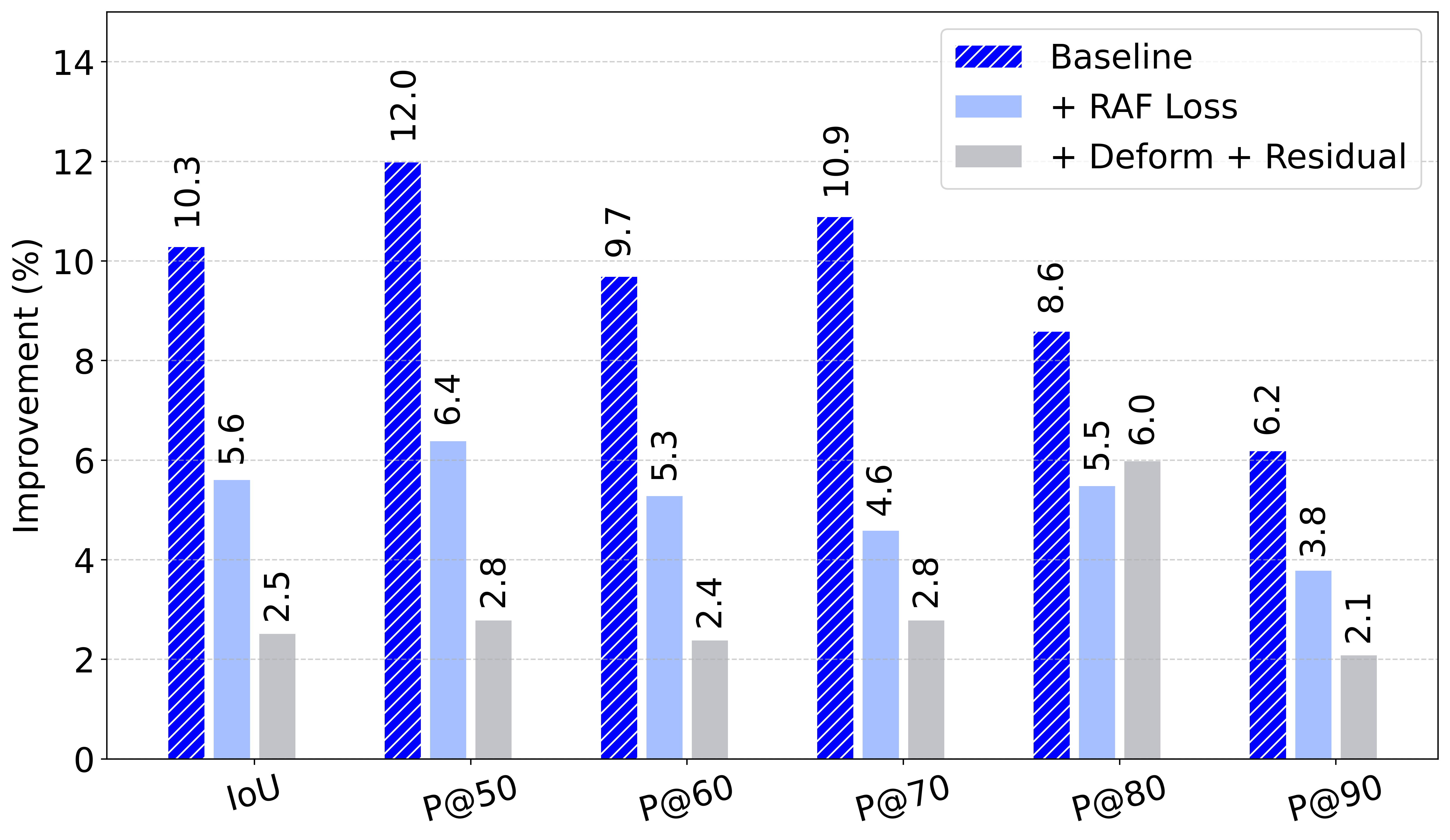}
    \caption{Incremental improvements in IoU and precision scores (P@50–P@90) at each stage of model enhancement. Each bar illustrates the relative performance gain introduced by a specific component (RAF loss, deformable, convolution, residual connections, and SE Block), reaching the final SegVLM performance.}
    \label{fig:ablation3}
\end{figure}

To evaluate the contribution of each proposed component, we conduct an ablation study in which modules in the baseline model are gradually replaced with our improvements and evaluated on the PhraseCut dataset.
As shown in Table \ref{tab:ablation2}, the inclusion of the RAF loss improves IoU by nearly 6\% and yields substantial gains at higher IoU thresholds. Further performance boosts are observed with the incorporation of deformable convolutions and residual connections which are particularly evident at Prec@90, thereby emphasizing enhanced boundary sensitivity. The integration of the SE block leads to the best overall performance across all metrics, confirming the complementary effect of each proposed enhancement in our proposed SegVLM model.

More importantly, we also show extended experimental results using bar charts in Fig. \ref{fig:ablation1},  Fig. \ref{fig:ablation2}, and  Fig.~\ref{fig:ablation3}. More specifically, in Fig. \ref{fig:ablation1}, we compare the incremental contributions of each proposed enhancement on segmentation precision across varying IoU thresholds. The results demonstrate a consistent performance improvement at each stage particularly under stricter evaluation criteria (e.g., IoU@90). Similarly, Fig. \ref{fig:ablation2} presents a comparison against state-of-the-art models, where SegVLM achieves the highest precision across all IoU thresholds, with especially notable improvements at higher thresholds (Prec@80 and Prec@90), highlighting its superior boundary alignment and robustness. Finally, Fig.~\ref{fig:ablation3} quantifies the relative performance gain contributed by each enhancement module, demonstrating that all components yield substantial improvements in both IoU and precision metrics, with the most significant gains observed at lower IoU thresholds (e.g., Prec@50).

\subsection{Qualitative Results}
\begin{figure*}[!t]
    \centering
\includegraphics[width=1.\linewidth]{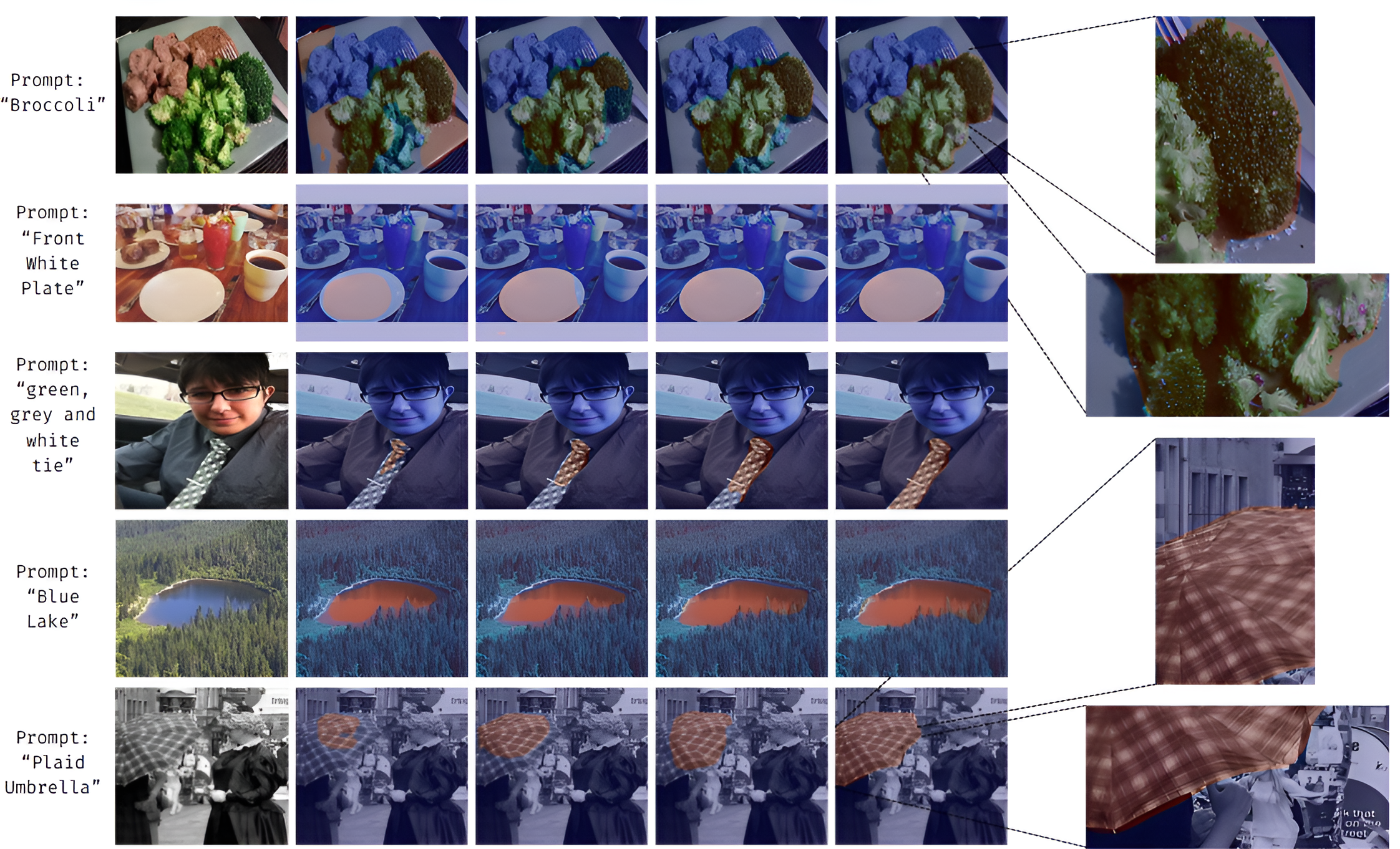}
    \caption{Qualitative evaluation of segmentation results across different model configurations. From left to right: (a) input image, (b) baseline output, followed by incremental integration of (c) RAF Loss, (d) deformable convolutions with residual connections, and (e) SE block. Each successive enhancement leads to improved localization accuracy and boundary refinement in the segmentation outputs.}
    \label{fig:qualitative_1}
\end{figure*}

In addition to quantitative evaluations, we provide qualitative examples that illustrate the visual impact of each proposed improvement. In Fig. \ref{fig:qualitative_1}, we show segmentation results generated by our model, demonstrating increased localization accuracy and alignment with the referring expressions across multiple settings. As a result, the baseline model frequently produces coarse or incomplete predictions, notably in situations requiring fine-grained classifications or visually ambiguous scenes. Integrating RAF loss increases the model’s sensitivity to hard examples and sharpens the mask. Additionally, we also examined the generalization of our model on datasets that are not dedicated for referring expression segmentation, such as the PASCAL VOC \cite{everingham2010pascal} and the MS COCO \cite{lin2014coco}. These datasets do not provide native prompts as RefCOCO or PhraseCut, so we manually created referring expressions for the objects in each image. 

\begin{figure*}[!t]
    \centering
\includegraphics[width=1.\linewidth]{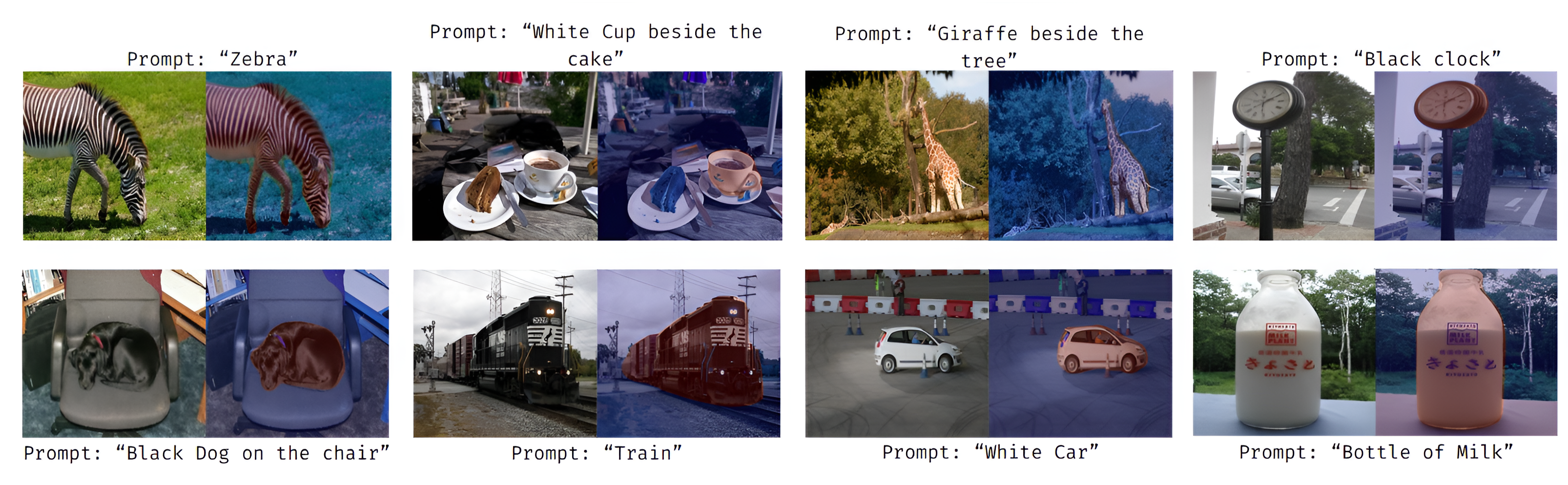}
    \caption{Qualitative results demonstrating the generalization capability of SegVLM on prompt-free datasets. The first row shows predictions on unseen images from the COCO dataset, while the second row presents results on the VOC dataset. In both cases, the model accurately distinguishes and segments target objects, highlighting its robustness across diverse visual domains.}
    \label{fig:fig:qualitative_2}
\end{figure*}

\begin{figure}[!t]
    \centering
\includegraphics[width=\linewidth]{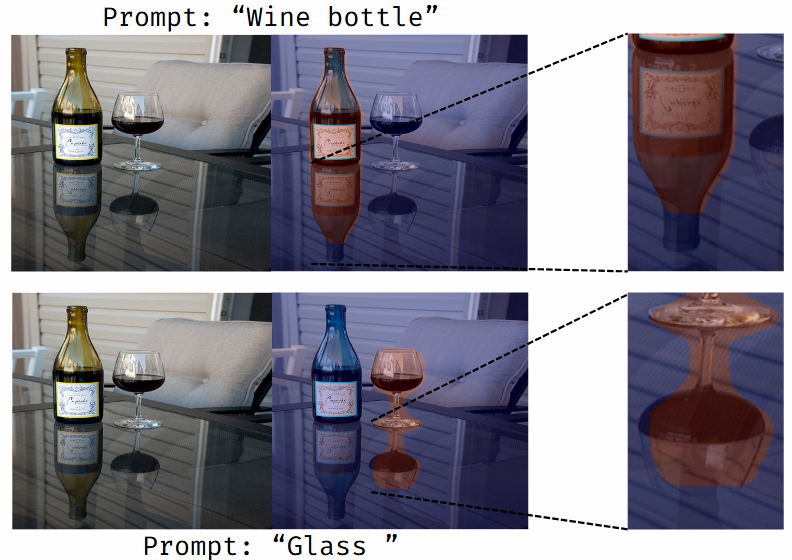}
    \caption{Segmentation result demonstrating the model's ability to capture fine-grained visual details, including reflections of the wine bottle and glass on the table. This highlights the effectiveness of the fine-tuned projection module and visual-language matching strategy in reasoning over contextual and spatial details, extending the model's capabilities beyond standard referring segmentation tasks.}
    \label{fig:fig:qualitative_3}
\end{figure}

In Fig. \ref{fig:fig:qualitative_2}, we show that our model generalizes well to segmentation across diverse categories and scenes when augmented with synthetic prompts, as demonstrated by the results in the top row of this table for the COCO dataset and the bottom row for the VOC dataset. This reflects the robustness of our model's multimodal understanding of. Also, during the generalization tests, we observed an emergent capability that our model can segment reflections and shadows. More importantly we show in Fig. \ref{fig:fig:qualitative_3} that the reflection of both the wine bottle and the glass on the table has been also segmented successfully . This demonstrates that our proposed project module and visual-language alignment  strategy allowed the model to reason over contextual and spatial details, including non-frontal object manifestations (e.g., mirrors). This behavior demonstrates the capability of the proposed architecture to handle broader visual reasoning tasks beyond standard referring segmentation.

\section{Conclusion}
In this work, we proposed SegVLM, a vision-language segmentation model for referring image segmentation. By integrating referring-aware fusion (RAF) loss, deformable convolutions with residual connections, and SE blocks, SegVLM achieves consistent improvements in IoU and precision across all thresholds, particularly at stricter levels. Despite its compact size (63M parameters), it outperforms several state-of-the-art baselines, demonstrating strong generalization across diverse scenes and categories. Our results highlight the model’s enhanced boundary sensitivity, robustness to visual ambiguity, and potential for broader visual reasoning tasks.

{
\bibliographystyle{elsarticle-num}
\bibliography{refs}

\begin{thebibliography}{10}
\expandafter\ifx\csname url\endcsname\relax
  \def\url#1{\texttt{#1}}\fi
\expandafter\ifx\csname urlprefix\endcsname\relax\def\urlprefix{URL }\fi
\expandafter\ifx\csname href\endcsname\relax
  \def\href#1#2{#2} \def\path#1{#1}\fi

\bibitem{long2015fcn}
J.~Long, E.~Shelhamer, T.~Darrell, Fully convolutional networks for semantic segmentation, in: Proceedings of the IEEE conference on computer vision and pattern recognition, 2015, pp. 3431--3440.

\bibitem{ma2024medsam}
J.~Ma, Y.~He, F.~Li, L.~Han, C.~You, B.~Wang, \href{https://www.nature.com/articles/s41467-024-44824-z}{Segment anything in medical images}, Nature Communications 15 (2024) 654.
\newblock \href {https://doi.org/10.1038/s41467-024-44824-z} {\path{doi:10.1038/s41467-024-44824-z}}.
\newline\urlprefix\url{https://www.nature.com/articles/s41467-024-44824-z}

\bibitem{hao2024real}
W.~Hao, J.~Wang, H.~Lu, \href{https://www.techscience.com/cmc/v81n3/59019}{A real-time semantic segmentation method based on transformer for autonomous driving}, Computers, Materials \& Continua 81~(3) (2024) 4419--4433.
\newblock \href {https://doi.org/10.32604/cmc.2024.055478} {\path{doi:10.32604/cmc.2024.055478}}.
\newline\urlprefix\url{https://www.techscience.com/cmc/v81n3/59019}

\bibitem{yu2024semantic}
C.-C. Yu, Y.-D. Chen, H.-Y. Cheng, C.-L. Jiang, \href{https://www.mdpi.com/1424-8220/24/20/6539}{Semantic segmentation of satellite images for landslide detection using foreground-aware and multi-scale convolutional attention mechanism}, Sensors 24~(20) (2024) 6539.
\newblock \href {https://doi.org/10.3390/s24206539} {\path{doi:10.3390/s24206539}}.
\newline\urlprefix\url{https://www.mdpi.com/1424-8220/24/20/6539}

\bibitem{zhang2024augmented}
X.~Zhang, Y.~Zhu, L.~Chen, P.~Duan, M.~Zhou, \href{https://www.nature.com/articles/s41598-024-65204-z}{Augmented reality navigation method based on image segmentation and sensor tracking registration technology}, Scientific Reports 14 (2024) 15281.
\newblock \href {https://doi.org/10.1038/s41598-024-65204-z} {\path{doi:10.1038/s41598-024-65204-z}}.
\newline\urlprefix\url{https://www.nature.com/articles/s41598-024-65204-z}

\bibitem{chen2017deeplab}
L.-C. Chen, G.~Papandreou, I.~Kokkinos, K.~Murphy, A.~L. Yuille, Deeplab: Semantic image segmentation with deep convolutional nets, atrous convolution, and fully connected crfs, IEEE Transactions on Pattern Analysis and Machine Intelligence 40~(4) (2017) 834--848.

\bibitem{liu2023gres}
C.~Liu, H.~Ding, X.~Jiang, Gres: Generalized referring expression segmentation, in: Proceedings of the IEEE/CVF Conference on Computer Vision and Pattern Recognition (CVPR), 2023.

\bibitem{yan2023mmnet}
T.~Yan, Y.~Xu, Z.~Zhao, H.~Wang, W.~Xie, Y.~Liu, Y.~Tian, Mmnet: Multi-mask network for referring image segmentation, arXiv preprint arXiv:2305.14969 (2023).

\bibitem{li2021mail}
Z.~Li, M.~Wang, J.~Mei, Y.~Liu, Mail: A unified mask-image-language trimodal network for referring image segmentation, arXiv preprint arXiv:2111.10747 (2021).

\bibitem{ding2021vision}
H.~Ding, C.~Liu, S.~Wang, X.~Jiang, Vision-language transformer and query generation for referring segmentation, arXiv preprint arXiv:2108.05565 (2021).

\bibitem{cho2024metris}
Y.~Cho, H.~Yu, K.~Kong, S.-J. Kang, Metris: Multi-expressions for transformer-based referring image segmentation, in: International Conference on Learning Representations (ICLR), 2025.

\bibitem{xiao2024oneref}
L.~Xiao, X.~Yang, F.~Peng, Y.~Wang, C.~Xu, Oneref: Unified one-tower expression grounding and segmentation with mask referring modeling, arXiv preprint arXiv:2410.08021 (2024).

\bibitem{reza2023mmsformer}
M.~K. Reza, A.~Prater-Bennette, M.~S. Asif, Mmsformer: Multimodal transformer for material and semantic segmentation, IEEE Open Journal of Signal Processing 4 (2023) 123--135.
\newblock \href {https://doi.org/10.1109/OJSP.2023.3291234} {\path{doi:10.1109/OJSP.2023.3291234}}.

\bibitem{radford2021learning}
A.~Radford, J.~W. Kim, C.~Hallacy, A.~Ramesh, G.~Goh, S.~Agarwal, G.~Sastry, A.~Askell, P.~Mishkin, J.~Clark, G.~Krueger, I.~Sutskever, \href{https://arxiv.org/abs/2103.00020}{Learning transferable visual models from natural language supervision}, in: Proceedings of the 38th International Conference on Machine Learning, PMLR, 2021.
\newline\urlprefix\url{https://arxiv.org/abs/2103.00020}

\bibitem{hu2016segmentation}
R.~Hu, M.~Rohrbach, T.~Darrell, Segmentation from natural language expressions, in: European Conference on Computer Vision (ECCV), Springer, 2016, pp. 108--124.

\bibitem{ruan2023seem}
T.~Ruan, Y.~Wang, Z.~Zhang, Y.~Li, \href{https://arxiv.org/abs/2304.06718}{Seem: Segment everything everywhere all at once}, arXiv preprint arXiv:2304.06718 (2023).
\newline\urlprefix\url{https://arxiv.org/abs/2304.06718}

\bibitem{li2022blip}
J.~Li, D.~Li, C.~Xiong, S.~C.~H. Hoi, \href{https://arxiv.org/abs/2201.12086}{Blip: Bootstrapping language-image pre-training for unified vision-language understanding and generation}, in: Proceedings of the 39th International Conference on Machine Learning (ICML), 2022.
\newline\urlprefix\url{https://arxiv.org/abs/2201.12086}

\bibitem{wang2022cris}
T.~Wang, X.~Zhang, H.~Zhao, Cris: Clip-driven referring image segmentation, in: Proceedings of the IEEE/CVF Conference on Computer Vision and Pattern Recognition, 2022, pp. 11686--11695.

\bibitem{yu2016modeling}
L.~Yu, P.~Poirson, S.~Yang, A.~C. Berg, T.~L. Berg, Modeling context in referring expressions, in: European Conference on Computer Vision (ECCV), 2016.

\bibitem{mao2016generation}
J.~Mao, J.~Huang, A.~Toshev, O.~Camburu, A.~Yuille, K.~Murphy, Generation and comprehension of unambiguous object descriptions, in: Proceedings of the IEEE Conference on Computer Vision and Pattern Recognition (CVPR), 2016.

\bibitem{wu2020phrasecut}
Z.~Wu, L.~Wang, Z.~Jiang, Y.~Xiong, D.~Lin, Phrasecut: Language-based image segmentation in the wild, in: Proceedings of the IEEE/CVF Conference on Computer Vision and Pattern Recognition, 2020, pp. 10219--10229.

\bibitem{liu2023referring}
Z.~Liu, et~al., Referring image segmentation using text supervision, in: Proceedings of the IEEE/CVF International Conference on Computer Vision (ICCV), IEEE, 2023, pp. 12345--12354.

\bibitem{alalyani2024multimodal}
R.~Alalyani, N.~Krishnaswamy, Multimodal referring expression generation in interactive virtual agents, in: International Conference on Human-Computer Interaction, Springer, 2024, pp. 3--22.

\bibitem{alalyani2024scmre}
R.~Alalyani, N.~Krishnaswamy, Scmre: A corpus for generative ai in multimodal human-computer interaction, in: International Conference on Human-Computer Interaction, Springer, 2024, pp. 23--39.

\bibitem{Tang_2023_CVPR}
J.~Tang, G.~Zheng, C.~Shi, S.~Yang, \href{https://openaccess.thecvf.com/content/CVPR2023/papers/Tang_Contrastive_Grouping_With_Transformer_for_Referring_Image_Segmentation_CVPR_2023_paper.pdf}{Contrastive grouping with transformer for referring image segmentation}, in: Proceedings of the IEEE/CVF Conference on Computer Vision and Pattern Recognition (CVPR), 2023, pp. 23570--23580.
\newblock \href {https://doi.org/10.1109/CVPR52729.2023.02257} {\path{doi:10.1109/CVPR52729.2023.02257}}.
\newline\urlprefix\url{https://openaccess.thecvf.com/content/CVPR2023/papers/Tang_Contrastive_Grouping_With_Transformer_for_Referring_Image_Segmentation_CVPR_2023_paper.pdf}

\bibitem{shridhar2018interactive}
M.~Shridhar, D.~Hsu, Interactive visual grounding of referring expressions for human-robot interaction, in: Proceedings of Robotics: Science and Systems (RSS), 2018.

\bibitem{paul2020grounding}
R.~Paul, R.~C. Arkin, S.~Chernova, Grounding spatio-semantic referring expressions for human-robot interaction, in: 2020 IEEE/RSJ International Conference on Intelligent Robots and Systems (IROS), IEEE, 2020, pp. 2328--2335.

\bibitem{yu2018mattnet}
L.~Yu, Z.~Lin, X.~Shen, J.~Yang, X.~Lu, M.~Bansal, T.~L. Berg, Mattnet: Modular attention network for referring expression comprehension, in: Proceedings of the IEEE Conference on Computer Vision and Pattern Recognition (CVPR), 2018, pp. 1307--1315.

\bibitem{NEURIPS2023_6dcf277e}
H.~Liu, C.~Li, Q.~Wu, Y.~J. Lee, \href{https://proceedings.neurips.cc/paper_files/paper/2023/file/6dcf277ea32ce3288914faf369fe6de0-Paper-Conference.pdf}{Visual instruction tuning}, in: A.~Oh, T.~Naumann, A.~Globerson, K.~Saenko, M.~Hardt, S.~Levine (Eds.), Advances in Neural Information Processing Systems, Vol.~36, Curran Associates, Inc., 2023, pp. 34892--34916.
\newline\urlprefix\url{https://proceedings.neurips.cc/paper_files/paper/2023/file/6dcf277ea32ce3288914faf369fe6de0-Paper-Conference.pdf}

\bibitem{NEURIPS2024_9ee3a664}
S.~Tong, E.~Brown, P.~Wu, S.~Woo, M.~Middepogu, S.~C. Akula, J.~Yang, S.~Yang, A.~Iyer, X.~Pan, A.~Wang, R.~Fergus, Y.~LeCun, S.~Xie, \href{https://proceedings.neurips.cc/paper_files/paper/2024/file/9ee3a664ccfeabc0da16ac6f1f1cfe59-Paper-Conference.pdf}{Cambrian-1: A fully open, vision-centric exploration of multimodal llms}, in: A.~Globerson, L.~Mackey, D.~Belgrave, A.~Fan, U.~Paquet, J.~Tomczak, C.~Zhang (Eds.), Advances in Neural Information Processing Systems, Vol.~37, Curran Associates, Inc., 2024, pp. 87310--87356.
\newline\urlprefix\url{https://proceedings.neurips.cc/paper_files/paper/2024/file/9ee3a664ccfeabc0da16ac6f1f1cfe59-Paper-Conference.pdf}

\bibitem{NEURIPS2024_03738e5f}
K.~Yan, Z.~Wang, L.~Ji, Y.~Wang, N.~Duan, S.~Ma, \href{https://proceedings.neurips.cc/paper_files/paper/2024/file/03738e5f26967582eeb3b57eef82f1f0-Paper-Conference.pdf}{Voila-a: Aligning vision-language models with user\textquotesingle s gaze attention}, in: A.~Globerson, L.~Mackey, D.~Belgrave, A.~Fan, U.~Paquet, J.~Tomczak, C.~Zhang (Eds.), Advances in Neural Information Processing Systems, Vol.~37, Curran Associates, Inc., 2024, pp. 1890--1918.
\newline\urlprefix\url{https://proceedings.neurips.cc/paper_files/paper/2024/file/03738e5f26967582eeb3b57eef82f1f0-Paper-Conference.pdf}

\bibitem{pmlr-v139-kim21k}
W.~Kim, B.~Son, I.~Kim, \href{https://proceedings.mlr.press/v139/kim21k.html}{Vilt: Vision-and-language transformer without convolution or region supervision}, in: M.~Meila, T.~Zhang (Eds.), Proceedings of the 38th International Conference on Machine Learning, Vol. 139 of Proceedings of Machine Learning Research, PMLR, 2021, pp. 5583--5594.
\newline\urlprefix\url{https://proceedings.mlr.press/v139/kim21k.html}

\bibitem{pmlr-v139-radford21a}
A.~Radford, J.~W. Kim, C.~Hallacy, A.~Ramesh, G.~Goh, S.~Agarwal, G.~Sastry, A.~Askell, P.~Mishkin, J.~Clark, G.~Krueger, I.~Sutskever, \href{https://proceedings.mlr.press/v139/radford21a.html}{Learning transferable visual models from natural language supervision}, in: M.~Meila, T.~Zhang (Eds.), Proceedings of the 38th International Conference on Machine Learning, Vol. 139 of Proceedings of Machine Learning Research, PMLR, 2021, pp. 8748--8763.
\newline\urlprefix\url{https://proceedings.mlr.press/v139/radford21a.html}

\bibitem{yang2022unitab}
Z.~Yang, Z.~Gan, J.~Wang, X.~Hu, F.~Ahmed, Z.~Liu, Y.~Lu, L.~Wang, Unitab: Unifying text and box outputs for grounded vision-language modeling, in: European Conference on Computer Vision (ECCV), Springer, 2022, pp. 521--539.

\bibitem{wang2022image}
W.~Wang, H.~Bao, L.~Dong, J.~Bjorck, Z.~Peng, Q.~Liu, K.~Aggarwal, O.~K. Mohammed, S.~Singhal, S.~Som, F.~Wei, Image as a foreign language: Beit pretraining for all vision and vision-language tasks, arXiv preprint arXiv:2208.10442 (2022).

\bibitem{bai2023qwen}
J.~Bai, S.~Bai, S.~Yang, S.~Wang, S.~Tan, P.~Wang, J.~Lin, C.~Zhou, J.~Zhou, Qwen-vl: A versatile vision-language model for understanding, localization, text reading, and beyond, arXiv preprint arXiv:2308.12966 (2023).

\bibitem{pmlr-v202-li23q}
J.~Li, D.~Li, S.~Savarese, S.~Hoi, \href{https://proceedings.mlr.press/v202/li23q.html}{{BLIP}-2: Bootstrapping language-image pre-training with frozen image encoders and large language models}, in: A.~Krause, E.~Brunskill, K.~Cho, B.~Engelhardt, S.~Sabato, J.~Scarlett (Eds.), Proceedings of the 40th International Conference on Machine Learning, Vol. 202 of Proceedings of Machine Learning Research, PMLR, 2023, pp. 19730--19742.
\newline\urlprefix\url{https://proceedings.mlr.press/v202/li23q.html}

\bibitem{zhu2024minigpt}
D.~Zhu, J.~Chen, X.~Shen, X.~Li, M.~Elhoseiny, \href{https://openreview.net/forum?id=1tZbq88f27}{Minigpt-4: Enhancing vision-language understanding with advanced large language models}, in: Proceedings of the International Conference on Learning Representations (ICLR), 2024.
\newline\urlprefix\url{https://openreview.net/forum?id=1tZbq88f27}

\bibitem{radford2021clip}
A.~Radford, J.~W. Kim, C.~Hallacy, A.~Ramesh, G.~Goh, S.~Agarwal, G.~Sastry, A.~Askell, P.~Mishkin, J.~Clark, G.~Krueger, I.~Sutskever, Learning transferable visual models from natural language supervision, in: Proceedings of the 38th International Conference on Machine Learning (ICML), 2021.

\bibitem{10.1145/3581783.3612117}
S.-A. Liu, Y.~Zhang, Z.~Qiu, H.~Xie, Y.~Zhang, T.~Yao, \href{https://doi.org/10.1145/3581783.3612117}{Caris: Context-aware referring image segmentation}, in: Proceedings of the 31st ACM International Conference on Multimedia, MM '23, Association for Computing Machinery, New York, NY, USA, 2023, p. 779–788.
\newblock \href {https://doi.org/10.1145/3581783.3612117} {\path{doi:10.1145/3581783.3612117}}.
\newline\urlprefix\url{https://doi.org/10.1145/3581783.3612117}

\bibitem{lads2023aaai}
A.~not specified, Referring expression comprehension using language adaptive dynamic subnetworks, in: Proceedings of the AAAI Conference on Artificial Intelligence, Vol.~37, 2023, pp. 13780--13800.

\bibitem{hemanthage2024recantformer}
B.~Hemanthage, H.~Bilen, P.~Bartie, C.~Dondrup, O.~Lemon, Recantformer: Referring expression comprehension with varying numbers of targets, in: Proceedings of the 2024 Conference on Empirical Methods in Natural Language Processing (EMNLP), Association for Computational Linguistics, 2024.

\bibitem{yang2023videococa}
G.~Yang, J.~Li, Y.~S. Wang, W.~Zhang, S.~C.~H. Hoi, Videococa: Video contrastive captioners are temporal learners, in: Advances in Neural Information Processing Systems (NeurIPS), 2023.

\bibitem{mei2024slvp}
J.~Mei, A.~Piergiovanni, J.-N. Hwang, W.~Li, Slvp: Self-supervised language-video pre-training for referring video object segmentation, in: Proceedings of the IEEE/CVF Winter Conference on Applications of Computer Vision Workshops (WACV Workshops), 2024.

\bibitem{zou2023xdecoder}
X.~Zou, Z.-Y. Dou, J.~Yang, Z.~Gan, L.~Li, C.~Li, X.~Dai, H.~Behl, J.~Wang, L.~Yuan, N.~Peng, L.~Wang, Y.~J. Lee, J.~Gao, Generalized decoding for pixel, image, and language, in: Proceedings of the IEEE/CVF Conference on Computer Vision and Pattern Recognition (CVPR), 2023.

\bibitem{wang2022omnivl}
J.~Wang, D.~Chen, Z.~Wu, C.~Luo, L.~Zhou, Y.~Zhao, Y.~Xie, C.~Liu, Y.-G. Jiang, L.~Yuan, Omnivl: One foundation model for image-language and video-language tasks, in: Advances in Neural Information Processing Systems (NeurIPS), 2022.

\bibitem{dai2023instructblip}
W.~Dai, J.~Li, D.~Li, A.~M.~H. Tiong, J.~Zhao, W.~Wang, B.~Li, P.~Fung, S.~Hoi, Instructblip: Towards general-purpose vision-language models with instruction tuning, in: Advances in Neural Information Processing Systems (NeurIPS), 2023.

\bibitem{1640964}
R.~Hadsell, S.~Chopra, Y.~LeCun, Dimensionality reduction by learning an invariant mapping, in: 2006 IEEE Computer Society Conference on Computer Vision and Pattern Recognition (CVPR'06), Vol.~2, 2006, pp. 1735--1742.
\newblock \href {https://doi.org/10.1109/CVPR.2006.100} {\path{doi:10.1109/CVPR.2006.100}}.

\bibitem{pmlr-v119-chen20j}
T.~Chen, S.~Kornblith, M.~Norouzi, G.~Hinton, \href{https://proceedings.mlr.press/v119/chen20j.html}{A simple framework for contrastive learning of visual representations}, in: H.~D. III, A.~Singh (Eds.), Proceedings of the 37th International Conference on Machine Learning, Vol. 119 of Proceedings of Machine Learning Research, PMLR, 2020, pp. 1597--1607.
\newline\urlprefix\url{https://proceedings.mlr.press/v119/chen20j.html}

\bibitem{10.1007/978-3-031-19809-0_17}
Y.~Ci, C.~Lin, L.~Bai, W.~Ouyang, \href{https://doi.org/10.1007/978-3-031-19809-0_17}{Fast-moco: Boost momentum-based contrastive learning with combinatorial patches}, in: Computer Vision – ECCV 2022: 17th European Conference, Tel Aviv, Israel, October 23–27, 2022, Proceedings, Part XXVI, Springer-Verlag, Berlin, Heidelberg, 2022, p. 290–306.
\newblock \href {https://doi.org/10.1007/978-3-031-19809-0_17} {\path{doi:10.1007/978-3-031-19809-0_17}}.
\newline\urlprefix\url{https://doi.org/10.1007/978-3-031-19809-0_17}

\bibitem{he2020moco}
K.~He, H.~Fan, Y.~Wu, S.~Xie, R.~Girshick, Momentum contrast for unsupervised visual representation learning, in: Proceedings of the IEEE/CVF Conference on Computer Vision and Pattern Recognition (CVPR), IEEE, 2020, pp. 9729--9738.

\bibitem{Li_2022_CVPR}
S.~Li, X.~Xia, S.~Ge, T.~Liu, Selective-supervised contrastive learning with noisy labels, in: Proceedings of the IEEE/CVF Conference on Computer Vision and Pattern Recognition (CVPR), 2022, pp. 316--325.
\newblock \href {https://doi.org/10.1109/CVPR52688.2022.00037} {\path{doi:10.1109/CVPR52688.2022.00037}}.

\bibitem{10.1109/TMM.2023.3324588}
Y.~Zhang, C.~Liu, Y.~Zhou, W.~Wang, Q.~Ye, X.~Ji, \href{https://doi.org/10.1109/TMM.2023.3324588}{Beyond instance discrimination: Relation-aware contrastive self-supervised learning}, Trans. Multi. 26 (2024) 4628–4640.
\newblock \href {https://doi.org/10.1109/TMM.2023.3324588} {\path{doi:10.1109/TMM.2023.3324588}}.
\newline\urlprefix\url{https://doi.org/10.1109/TMM.2023.3324588}

\bibitem{10.1145/3581783.3612247}
J.~Zhang, T.~Lin, Y.~Xu, K.~Chen, R.~Zhang, \href{https://doi.org/10.1145/3581783.3612247}{Relational contrastive learning for scene text recognition}, in: Proceedings of the 31st ACM International Conference on Multimedia, MM '23, Association for Computing Machinery, New York, NY, USA, 2023, p. 5764–5775.
\newblock \href {https://doi.org/10.1145/3581783.3612247} {\path{doi:10.1145/3581783.3612247}}.
\newline\urlprefix\url{https://doi.org/10.1145/3581783.3612247}

\bibitem{pinheiro2020vader}
P.~O. Pinheiro, A.~Almahairi, R.~Y. Benmalek, F.~Golemo, A.~Courville, \href{https://arxiv.org/abs/2011.05499}{Unsupervised learning of dense visual representations}, in: Advances in Neural Information Processing Systems (NeurIPS), 2020.
\newline\urlprefix\url{https://arxiv.org/abs/2011.05499}

\bibitem{wang2021dense}
X.~Wang, R.~Zhang, C.~Shen, T.~Kong, L.~Li, Dense contrastive learning for self-supervised visual pre-training, in: Proceedings of the IEEE/CVF Conference on Computer Vision and Pattern Recognition (CVPR), 2021, pp. 3023--3032.
\newblock \href {https://doi.org/10.1109/CVPR46437.2021.00302} {\path{doi:10.1109/CVPR46437.2021.00302}}.

\bibitem{li2018referring}
R.~Li, K.~Li, Y.-C. Kuo, M.~Shu, X.~Qi, X.~Shen, J.~Jia, Referring image segmentation via recurrent refinement networks, in: Proceedings of the IEEE Conference on Computer Vision and Pattern Recognition (CVPR), 2018, pp. 5745--5753.

\bibitem{liu2017recurrent}
C.~Liu, Z.~Lin, X.~Shen, J.~Yang, X.~Lu, A.~Yuille, Recurrent multimodal interaction for referring image segmentation, in: Proceedings of the IEEE International Conference on Computer Vision (ICCV), 2017, pp. 1271--1280.

\bibitem{margffoy2018dynamic}
E.~Margffoy-Tuay, J.~C. Pérez, E.~Botero, P.~Arbeláez, Dynamic multimodal instance segmentation guided by natural language queries, in: Proceedings of the European Conference on Computer Vision (ECCV), Springer, 2018, pp. 630--645.

\bibitem{hochreiter1997long}
S.~Hochreiter, J.~Schmidhuber, Long short-term memory, Neural Computation 9~(8) (1997) 1735--1780.
\newblock \href {https://doi.org/10.1162/neco.1997.9.8.1735} {\path{doi:10.1162/neco.1997.9.8.1735}}.

\bibitem{long2015fully}
J.~Long, E.~Shelhamer, T.~Darrell, Fully convolutional networks for semantic segmentation, in: Proceedings of the IEEE Conference on Computer Vision and Pattern Recognition (CVPR), 2015, pp. 3431--3440.

\bibitem{he2017mask}
K.~He, G.~Gkioxari, P.~Dollár, R.~Girshick, Mask r-cnn, in: Proceedings of the IEEE International Conference on Computer Vision (ICCV), 2017, pp. 2961--2969.

\bibitem{luo2020multi}
G.~Luo, Y.~Zhou, X.~Sun, L.~Cao, C.~Wu, C.~Deng, R.~Ji, Multi-task collaborative network for joint referring expression comprehension and segmentation, in: Proceedings of the IEEE/CVF Conference on Computer Vision and Pattern Recognition (CVPR), 2020, pp. 10034--10043.

\bibitem{zhang2023risam}
M.~Zhang, Y.~Liu, X.~Yin, H.~Yue, J.~Yang, \href{https://arxiv.org/abs/2311.15727}{Risam: Referring image segmentation via mutual-aware attention features}, arXiv preprint arXiv:2311.15727 (2023).
\newline\urlprefix\url{https://arxiv.org/abs/2311.15727}

\bibitem{yang2022lavt}
Z.~Yang, J.~Wang, Y.~Tang, K.~Chen, H.~Zhao, P.~H. Torr, Lavt: Language-aware vision transformer for referring image segmentation, in: Proceedings of the IEEE/CVF Conference on Computer Vision and Pattern Recognition (CVPR), IEEE, 2022, pp. 17247--17256.
\newblock \href {https://doi.org/10.1109/CVPR52688.2022.01680} {\path{doi:10.1109/CVPR52688.2022.01680}}.

\bibitem{ouyang2023slvit}
S.~Ouyang, H.~Wang, S.~Xie, Z.~Niu, R.~Tong, Y.-W. Chen, L.~Lin, \href{https://www.ijcai.org/proceedings/2023/0144.pdf}{Slvit: Scale-wise language-guided vision transformer for referring image segmentation}, in: Proceedings of the 32nd International Joint Conference on Artificial Intelligence (IJCAI), International Joint Conferences on Artificial Intelligence Organization, 2023, pp. 1299--1306.
\newline\urlprefix\url{https://www.ijcai.org/proceedings/2023/0144.pdf}

\bibitem{ye2019cross}
L.~Ye, M.~Rochan, Z.~Liu, Y.~Wang, Cross-modal self-attention network for referring image segmentation, in: Proceedings of the IEEE/CVF Conference on Computer Vision and Pattern Recognition (CVPR), IEEE, 2019, pp. 10502--10511.

\bibitem{hu2020bi}
Z.~Hu, G.~Feng, J.~Sun, L.~Zhang, H.~Lu, Bi-directional relationship inferring network for referring image segmentation, in: Proceedings of the IEEE/CVF Conference on Computer Vision and Pattern Recognition (CVPR), IEEE, 2020, pp. 4424--4433.

\bibitem{yan2024fuse}
Y.~Yan, X.~He, S.~Chen, S.~Lu, J.~Liu, \href{https://arxiv.org/abs/2405.11205}{Fuse \& calibrate: A bi-directional vision-language guided framework for referring image segmentation}, arXiv preprint arXiv:2405.11205 (2024).
\newline\urlprefix\url{https://arxiv.org/abs/2405.11205}

\bibitem{shang2024prompt}
C.~Shang, Z.~Song, H.~Qiu, L.~Wang, F.~Meng, H.~Li, Prompt-guided bidirectional deep fusion network for referring image segmentation, Neurocomputing (2024).
\newblock \href {https://doi.org/10.1016/j.neucom.2024.01.123} {\path{doi:10.1016/j.neucom.2024.01.123}}.

\bibitem{dai2017deformable}
J.~Dai, H.~Qi, Y.~Xiong, Y.~Li, G.~Zhang, H.~Hu, Y.~Wei, Deformable convolutional networks, in: Proceedings of the IEEE International Conference on Computer Vision (ICCV), 2017, pp. 764--773.

\bibitem{goodfellow2016deep}
I.~Goodfellow, Y.~Bengio, A.~Courville, \href{https://www.deeplearningbook.org/}{Deep Learning}, MIT Press, 2016.
\newline\urlprefix\url{https://www.deeplearningbook.org/}

\bibitem{lin2017focal}
T.-Y. Lin, P.~Goyal, R.~Girshick, K.~He, P.~Doll{\'a}r, Focal loss for dense object detection, in: Proceedings of the IEEE international conference on computer vision, 2017, pp. 2980--2988.

\bibitem{milletari2016vnet}
F.~Milletari, N.~Navab, S.-A. Ahmadi, V-net: Fully convolutional neural networks for volumetric medical image segmentation, in: 2016 Fourth International Conference on 3D Vision (3DV), IEEE, 2016, pp. 565--571.

\bibitem{chu2019lmdb}
H.~Chu, \href{https://doi.org/10.1109/MS.2019.2936273}{Howard chu on lightning memory-mapped database}, IEEE Software 36~(6) (2019) 96--100.
\newblock \href {https://doi.org/10.1109/MS.2019.2936273} {\path{doi:10.1109/MS.2019.2936273}}.
\newline\urlprefix\url{https://doi.org/10.1109/MS.2019.2936273}

\bibitem{zhang2018context}
H.~Zhang, K.~Dana, J.~Shi, Z.~Zhang, X.~Wang, A.~Tyagi, A.~Agrawal, Context encoding for semantic segmentation, in: Proceedings of the IEEE Conference on Computer Vision and Pattern Recognition, 2018, pp. 7151--7160.

\bibitem{pascanu2013difficulty}
R.~Pascanu, T.~Mikolov, Y.~Bengio, On the difficulty of training recurrent neural networks, in: International Conference on Machine Learning (ICML), 2013, pp. 1310--1318.

\bibitem{he2015delving}
K.~He, X.~Zhang, S.~Ren, J.~Sun, Delving deep into rectifiers: Surpassing human-level performance on imagenet classification, in: Proceedings of the IEEE International Conference on Computer Vision (ICCV), 2015, pp. 1026--1034.

\bibitem{Kamath_2021_ICCV}
A.~Kamath, M.~Singh, Y.~LeCun, G.~Synnaeve, I.~Misra, N.~Carion, Mdetr -- modulated detection for end-to-end multi-modal understanding, in: Proceedings of the IEEE/CVF International Conference on Computer Vision (ICCV), 2021, pp. 1760--1770.

\bibitem{zhang2024groundhog}
Y.~Zhang, Z.~Ma, X.~Gao, S.~Shakiah, Q.~Gao, J.~Chai, Groundhog: Grounding large language models to holistic segmentation, in: Proceedings of the IEEE/CVF Conference on Computer Vision and Pattern Recognition (CVPR), 2024, pp. 14227--14238.
\newblock \href {https://doi.org/10.1109/CVPR52733.2024.01349} {\path{doi:10.1109/CVPR52733.2024.01349}}.

\bibitem{Zhang_2024_CVPR}
Y.~Zhang, Z.~Ma, X.~Gao, S.~Shakiah, Q.~Gao, J.~Chai, Groundhog: Grounding large language models to holistic segmentation, in: Proceedings of the IEEE/CVF Conference on Computer Vision and Pattern Recognition (CVPR), 2024, pp. 14227--14238.

\bibitem{everingham2010pascal}
M.~Everingham, L.~Van~Gool, C.~K. Williams, J.~Winn, A.~Zisserman, The pascal visual object classes (voc) challenge, International Journal of Computer Vision (IJCV) 88~(2) (2010) 303--338.

\bibitem{lin2014coco}
T.-Y. Lin, M.~Maire, S.~Belongie, J.~Hays, P.~Perona, D.~Ramanan, P.~Doll{\'a}r, C.~L. Zitnick, Microsoft coco: Common objects in context, in: European Conference on Computer Vision (ECCV), Springer, 2014, pp. 740--755.

\end{thebibliography}
}

\end{document}